\documentclass{article}
\usepackage[T1]{fontenc}
\usepackage{preprint,times}
\usepackage{amsmath,amssymb,bm}
\usepackage{booktabs,multirow,array,tabularx,graphicx,adjustbox,float}
\usepackage{xcolor,colortbl,microtype,xspace}
\usepackage{wrapfig}
\usepackage{hyperref,url}
\usepackage{enumitem}

\definecolor{citeblue}{RGB}{54,125,189}
\definecolor{refred}{RGB}{255,0,0}
\definecolor{urlrose}{RGB}{236,0,140}
\definecolor{tablepurple}{HTML}{EEE7F3}
\definecolor{tableblue}{HTML}{E6F0FA}

\hypersetup{
    colorlinks=true,
    citecolor=citeblue,
    linkcolor=refred,
    urlcolor=urlrose
}
\graphicspath{{figures/}}
\newcommand{\method}{\textsc{SimVTA}\xspace}
\newcommand{\epona}{\textsc{Epona}\xspace}

\newcommand{\Ltraj}{\mathcal{L}_{\mathrm{traj}}}
\newcommand{\Lvideo}{\mathcal{L}_{\mathrm{video}}}
\newcommand{\Lanchor}{\mathcal{L}_{\mathrm{anchor}}}
\newcommand{\sg}{\operatorname{sg}}
\title{\method: Simulation-Grounded Video-Trajectory Alignment for Driving World Action Models}
\author{%
  \begin{minipage}[t]{\dimexpr\textwidth-2\tabcolsep\relax}
    \normalfont\normalsize\bfseries
    Changhui Jing\textsuperscript{1,2,4}\quad
    Haochen Tian\textsuperscript{2$\dagger$}\quad
    Hongchen Li\textsuperscript{1,2,4}\quad
    Tianyu Li\textsuperscript{1}\quad
    Caojun Wang\textsuperscript{1}\quad \\
    Zihao Zhang\textsuperscript{2}\quad
    Yanhao Liu\textsuperscript{1}\quad
    Chengen Xie\textsuperscript{1}\quad
    Bailin Li\textsuperscript{3}\quad
    Hao Zhao\textsuperscript{5}\quad
    Zhenjie Yang\textsuperscript{2$\ddagger$}
    \\[8pt]
    \normalfont\small
    \textsuperscript{1}Shanghai Innovation Institute  \quad
    \textsuperscript{2}OpenDriveLab at The University of Hong Kong  \\
    \textsuperscript{3}Li Auto Inc.  \quad
    \textsuperscript{4}Tongji University \quad
    \textsuperscript{5}Tsinghua University \\[1mm]
    $^{\dagger}$Project Lead. $^{\ddagger}$Correspondence Author. \\[6pt]
    \href{https://opendrivelab.com/SimVTA/}{%
      \textcolor{urlrose}{\normalfont\normalsize\texttt{https://opendrivelab.com/SimVTA/}}%
    }
  \end{minipage}%
}
\iclrfinalcopy
\begin{document}
\maketitle
\fancyhead{}
\lhead{Preprint}
\renewcommand{\headrulewidth}{0.4pt}

\begin{abstract}
Driving world action models (WAMs) learn from paired trajectories and
future observations. Post-training with reinforcement learning allows the policy to explore alternative trajectories and improve planning using driving rewards. However, driving logs contain future videos only for the recorded trajectories, leaving these alternatives without corresponding visual supervision.
Thus, we introduce \method, a simulation-grounded post-training framework that combines trajectory optimization with video supervision. First, we sample candidate trajectories for each observation and use a real-world simulator (e.g., World Engine) to render future videos for high-reward candidates
in real-world scenes reconstructed with 3D Gaussian Splatting (3DGS).
This provides trajectory-matched supervision independent of the
model's own predictions.
Under a limited rendering budget, we prioritize video supervision
for higher-error futures among these rendered candidates. Finally, we optimize trajectory planning with Group Relative Policy
Optimization (GRPO) and use a flow-matching loss on the matched videos
to update shared representations. At inference time, the model predicts
trajectories directly, without simulator calls. On NAVSIM v1, applying \method to a WAM initialization
improves PDMS from 86.2 to 91.3 (+5.1).
It achieves 88.2 EPDMS on NAVSIM v2 and an average HD-Score
of 38.8 in closed-loop evaluation on HUGSIM. Ablations show that updating shared representations with video
supervision improves planning, while prioritizing higher-error futures
among high-reward trajectories further improves performance.
The results and demo are available at \url{https://github.com/OpenDriveLab/SimVTA}.
\end{abstract}

\begin{figure}[!htbp]
\centering
\includegraphics[width=.94\linewidth]{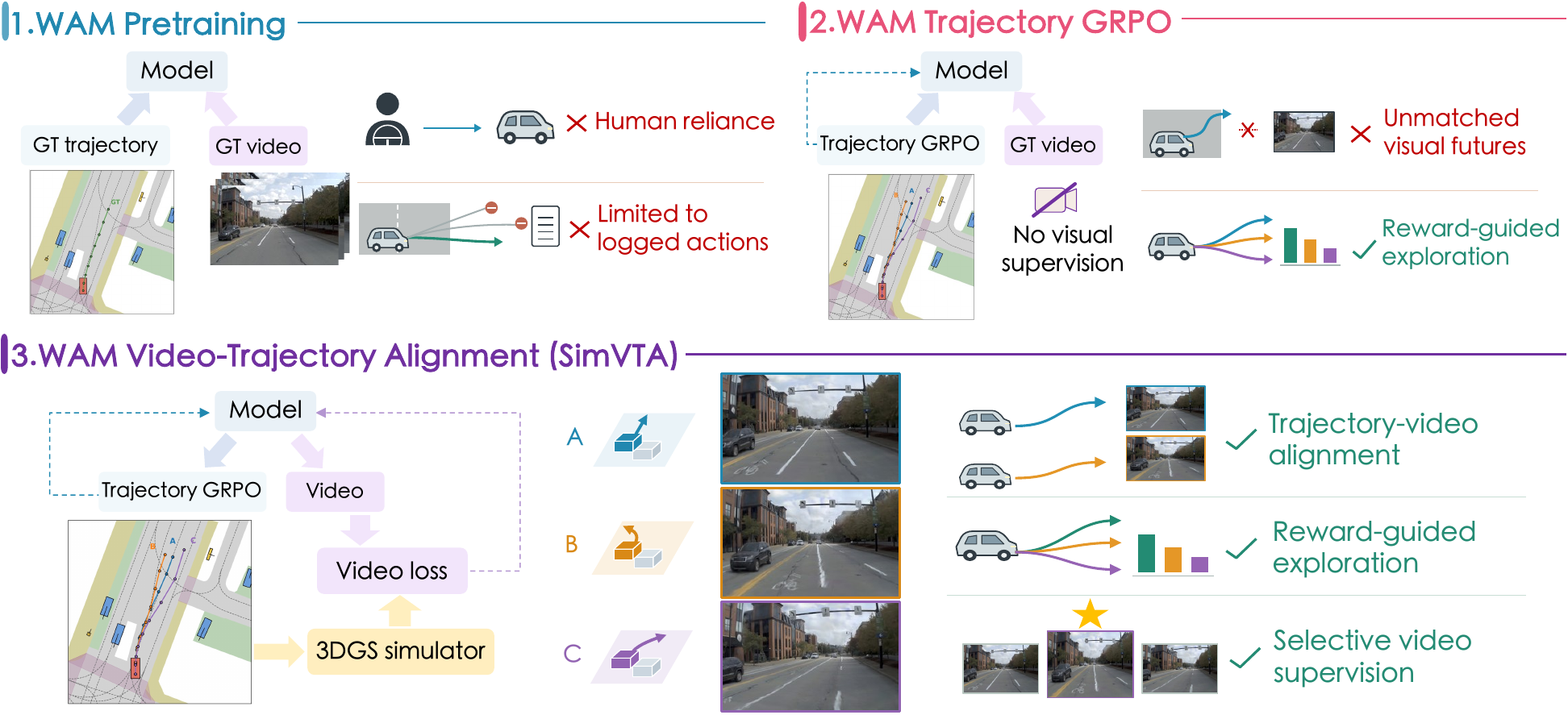}
\caption{\textbf{From logged futures to candidate-matched supervision.}
(1) WAM pretraining uses the ground-truth (GT) trajectory and its
logged future video.
(2) Trajectory GRPO explores candidate trajectories using driving
rewards, without matched video supervision.
(3) World Engine renders reconstructed scenes along selected candidate paths,
providing matched supervision independent of the model's predictions.}
\label{fig:motivation}
\end{figure}
\section{Introduction}
Driving WAMs combine future-scene prediction with
trajectory planning for the ego vehicle. Predicting future frames provides
dense visual supervision that complements sparse trajectory targets and
helps models learn representations for planning
\citep{li2025drivevla}. Recent driving models retain video prediction
during training but predict trajectories directly at inference time,
avoiding the computational cost of future-video generation
\citep{zhao2026simwam}.

Reinforcement learning can refine a planner trained on demonstrations to explore alternative trajectories and learn from driving
rewards \citep{zhou2025autovla}.
However, driving logs contain future videos only for recorded trajectories.
For example, when a candidate turns while the recorded trajectory continues
straight, the logged video does not show the visual consequences.
Reusing such videos creates incorrect trajectory-video pairs, while
omitting video training leaves these candidates without visual supervision.
This creates a supervision gap during post-training: current-policy
trajectories lack matched future-video targets
(Figure~\ref{fig:motivation}).

To address this gap, we introduce \method, a simulation-grounded
framework for video-trajectory alignment. A WAM's own video
predictions provide no external reference for checking whether they
follow a candidate trajectory. We use the 3DGS-based simulator in
World Engine \citep{li2026worldengine} to render reconstructed
real-world scenes from camera poses along executed candidate paths
(Figure~\ref{fig:motivation}). These renderings provide prediction-independent targets that match
the scene geometry and motion of each selected trajectory, establishing
\emph{video-trajectory alignment}.

We jointly train video prediction with flow matching and trajectory
planning with GRPO, allowing the video loss to update shared representations.
The former teaches the model to predict the visual consequences of
candidate trajectories; the latter compares driving rewards among
candidates and favors higher-reward
trajectories.

Rendering videos for every sampled trajectory is costly. We first
select high-reward candidates for rendering, then prioritize video
supervision for higher-error futures among the rendered candidates. This separates
trajectory exploration from video-supervision target selection.
Our ablations examine shared representation updates, target selection,
trajectory-video correspondence, and online versus offline rendering. At inference, the planner predicts
trajectories directly, without simulator calls or future-video generation.

Our main contributions are summarized as follows:
\begin{itemize}
[leftmargin=*,topsep=2pt,itemsep=1pt,parsep=0pt,partopsep=0pt]
    \item We introduce \method, a post-training framework that
combines trajectory GRPO with simulator-based video supervision.
It connects policy sampling, trajectory execution, rendering, and
joint updates to learn the visual consequences of actions beyond
those recorded in driving logs.
    \item We introduce a video-target selection strategy based on
trajectory reward and video prediction error. It prioritizes higher-error
futures of high-reward trajectories for video supervision without
restricting trajectory exploration.
    \item Applying \method to a WAM initialization improves
NAVSIM v1 PDMS by 5.1 points, from 86.2 to 91.3, and achieves
88.2 EPDMS on NAVSIM v2 and an average HD-Score of 38.8 on HUGSIM.
\end{itemize}
\begin{figure}[!htbp]
\centering
\includegraphics[width=1\linewidth]{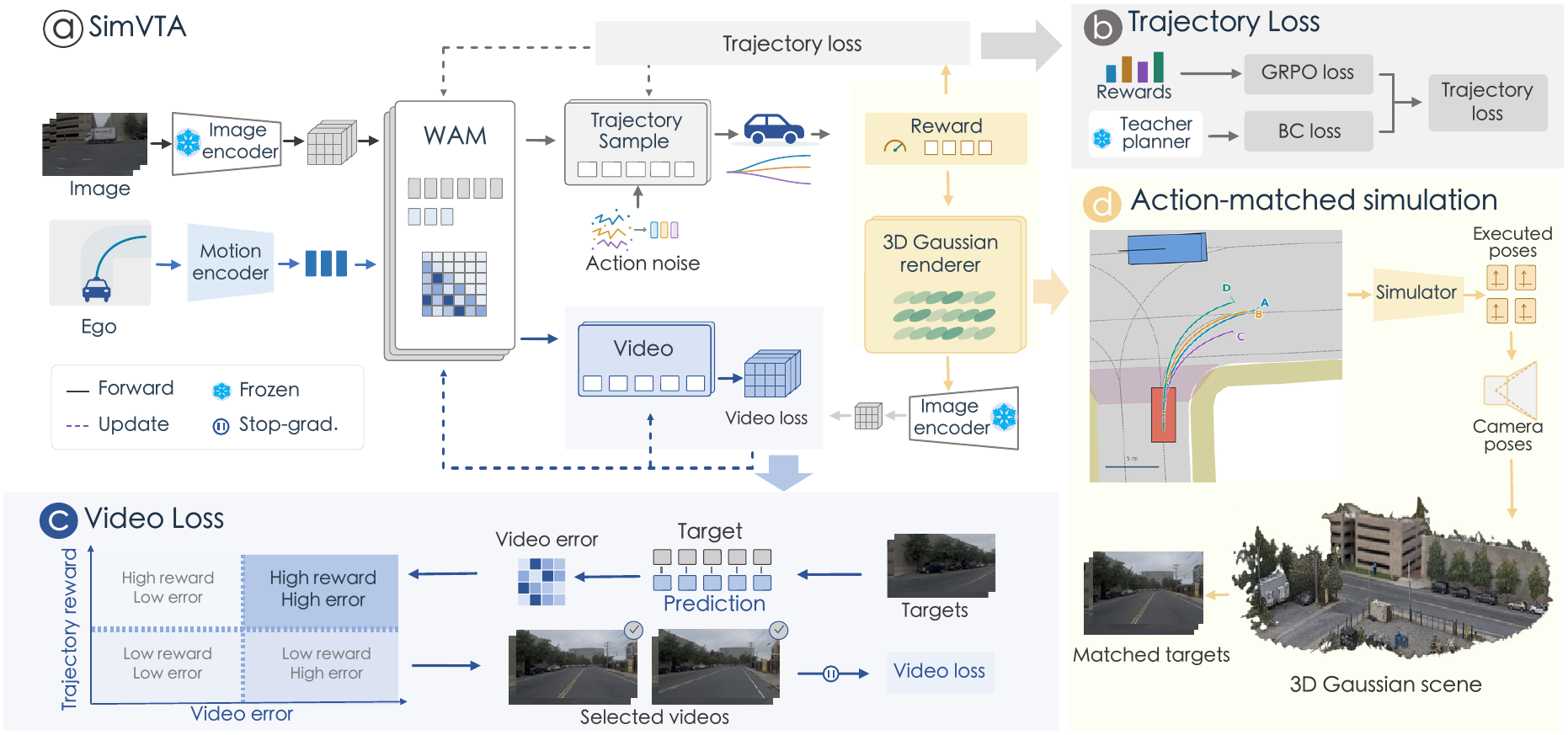}
\caption{\textbf{Simulation-grounded video--trajectory alignment.}
(a) Training overview.
(b) Trajectory learning combines GRPO with a behavior-cloning (BC) regularizer.
(c) Video supervision selects high-reward, higher-error futures.
(d) World Engine renders matched futures along executed paths.}
\label{fig:overview}
\end{figure}
\section{Related Work}
\label{sec:related-work}
\paragraph{World Action Models.}
World models predict future observations or latent states to support
driving simulation and planning~\citep{wang2024drivedreamer,
gao2024vista,li2025drivevla,yang2025raw2drive,wang2026latentwam,wang2026wajepa,
jia2026driveworldvla,lu2026onevl}.
Video-based approaches jointly model future frames and
trajectories~\citep{zhang2025epona,liu2026driveva,
shi2026drivewam,liu2026universe}.
Video prediction can also be retained during training and omitted
at inference, as explored in robotics by Fast-WAM
\citep{yuan2026fastwam} and in driving by SimWAM
\citep{zhao2026simwam}.
\method focuses on trajectory-matched video supervision
during driving WAM post-training.

\paragraph{Simulation for Driving.}
Driving simulators support policy training and evaluation through
controlled rollouts~\citep{dosovitskiy17carla,li2022metadrive,
zhou2024hugsim,cusumano2025selfplay,nvidia2025alpasim,
tian2026simscale,li2026worldengine,xiaomi2026drivezero}.
3DGS \citep{kerbl2023_3dgs} renders reconstructed scenes from
specified camera poses, providing visual observations for
closed-loop simulation~\citep{zhou2024hugsim,nvidia2025alpasim}.
World Engine \citep{li2026worldengine} renders these observations
along simulated ego motion for policy post-training.
CausalDrive \citep{yan2026causaldrive} instead uses a learned
renderer.
\method uses World Engine to render video targets along
selected candidate trajectories.

\paragraph{Post-Training for Driving.}
Driving policies can be refined after initial training through
reinforcement learning or model adaptation~\citep{wang2025adawm,zhang2025carplanner,
yang2025drivemoe,tang2026planr1,li2025recogdrive,zhou2025autovla,
nvidia2025alpamayo,li2026plannerrft}.
GRPO \citep{shao2024deepseekmath} has been applied to both
diffusion and autoregressive planners~\citep{li2025recogdrive,
tang2026planr1}.
PlannerRFT \citep{li2026plannerrft} uses closed-loop reinforcement
fine-tuning with adaptive exploration to improve diffusion planners.
Our trajectory optimization uses the stochastic differential
equation (SDE) sampling formulation from
Flow-GRPO \citep{liu2025flowgrpo}.
ReSim \citep{yang2025resim} trains a video simulator on simulated
non-expert trajectories to improve action controllability.
\method jointly updates shared representations through video
prediction and trajectory GRPO.

\begin{figure}[!htbp]
\centering
\includegraphics[width=\linewidth]{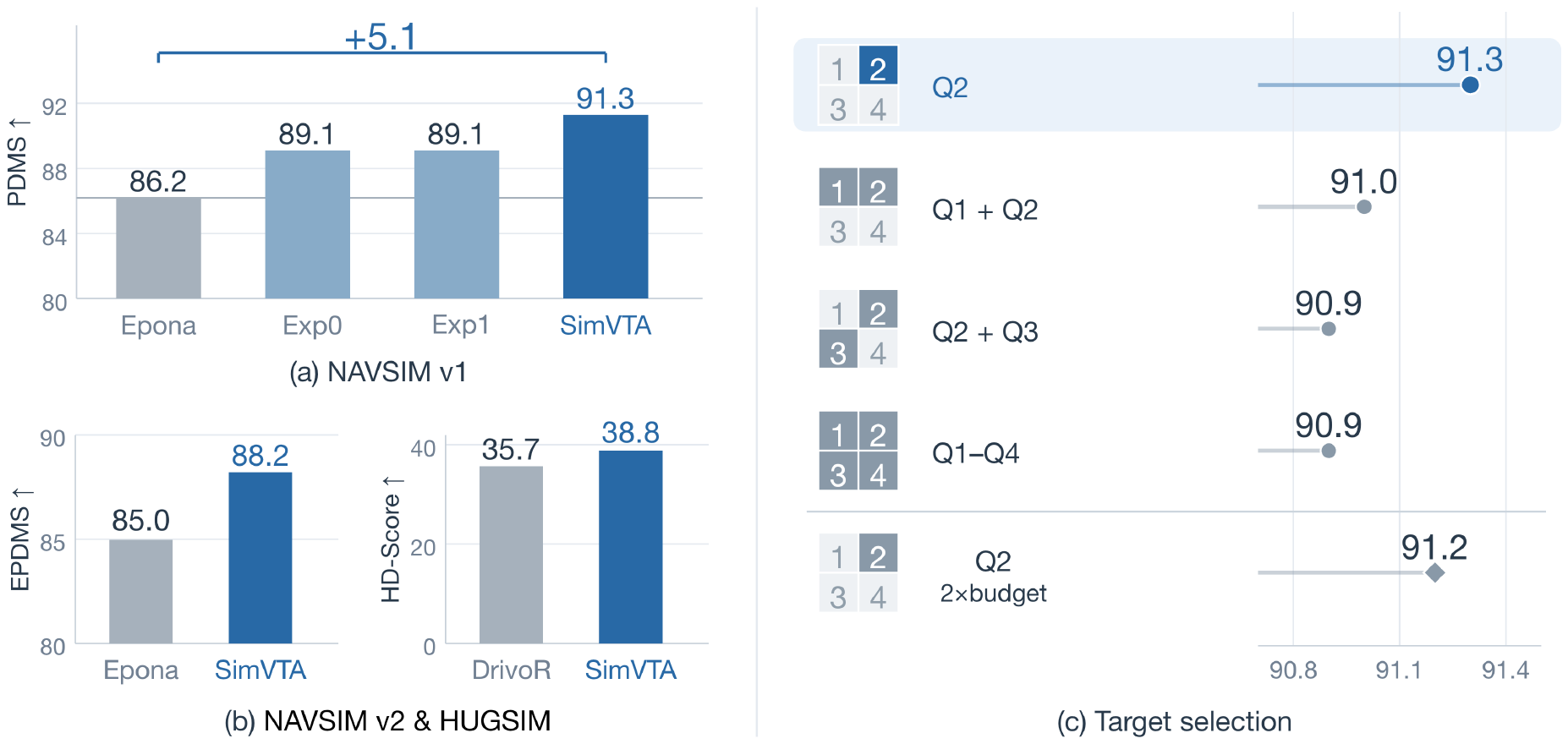}
\caption{\textbf{Planning results and video-supervision comparisons.}
(a) NAVSIM v1. (b) NAVSIM v2 and HUGSIM.
(c) Video-supervision configurations and rendering budget on NAVSIM v1.}
\label{fig:planning-gains}
\end{figure}
\section{Method}
\label{sec:method}

\method's post-training framework is illustrated in
Figure~\ref{fig:overview} and comprises trajectory sampling,
video rendering, and video supervision.
We sample candidate trajectories from the current policy and execute
them to obtain paths and driving rewards. World Engine
\citep{li2026worldengine} renders videos along selected paths during training. We select video targets based on
trajectory reward and video prediction error, then update the model with
trajectory GRPO and video supervision. The next sampling round uses
the updated policy.

\subsection{Trajectory Sampling}
Given observation history $o$, we compute a shared representation
$F=h_\theta(o)$ from visual and ego-motion history for trajectory and
video prediction.
The trajectory policy $\pi_\psi$ uses SDE sampling to generate
$G$ candidate trajectories:
\begin{equation}
 F=h_\theta(o),\qquad
 \tau_i\sim\pi_\psi(\cdot\mid F),\qquad i=1,\ldots,G.
 \label{eq:policy}
\end{equation}
NAVSIM's PDM executor $\mathcal{E}$ converts each candidate into an
executed ego path $\tilde\tau_i=\mathcal{E}(o,\tau_i)$.
We compute its reward $R_i$ from the PDM component scores,
using the progress-weighted aggregation in
Eq.~\ref{eq:training-reward}.
All $G$ candidates can contribute to trajectory learning.

\subsection{Video Rendering}
We first check candidate execution validity, motion-conditioning
support, and safety, then rank eligible candidates by $R_i$.
World Engine renders the highest-reward candidates within the
rendering budget $K$. These checks determine eligibility for
video supervision and do not remove candidates from trajectory learning.
For each selected candidate, the simulator $\mathcal{S}$ renders a target
video along its executed path, and the video model $W_\phi$ predicts the
future along the same path:
\begin{equation}
 V_i^{\rm sim}=\mathcal{S}(o,\tilde\tau_i),\qquad
 V_i^{\rm pred}=W_\phi(h_\theta(o),\tilde\tau_i).
 \label{eq:matched-target}
\end{equation}
Here $W_\phi$ denotes the video prediction process.
We render the reconstructed scene from camera poses along the
executed path, as shown in Figure~\ref{fig:overview}(d).
The same path conditions video prediction, so both the rendered
target and the prediction correspond to the same ego motion.

Rendering during training allows visual supervision to follow changes
in the policy. A fixed offline cache contains videos only for trajectories sampled
before optimization, whereas policy updates continuously change the
trajectory distribution explored during post-training. The stored videos remain valid
for their original paths, but do not provide matching targets for
new paths with different motions or viewpoints. In each sampling
round, \method renders videos for selected candidates from the current
policy, maintaining trajectory-video correspondence as training proceeds.

\subsection{Video Supervision}
\paragraph{Video Loss.}
Let $z^{\rm sim}_{i,r}$ be the image latent of future frame $r$ in
$V_i^{\rm sim}$, distinct from the shared representation $F$.
Flow matching \citep{lipman2023flowmatching,liu2022rectifiedflow}
interpolates between Gaussian noise and the target latent:
\begin{equation}
 x_{i,r,t}=t z^{\rm sim}_{i,r}+(1-t)\epsilon,\qquad
 u_{i,r}^\star=z^{\rm sim}_{i,r}-\epsilon,
 \quad t\sim\mathcal{U}(0,1),\;\epsilon\sim\mathcal{N}(0,I).
 \label{eq:flow-path}
\end{equation}
The velocity network $D_\phi$ conditions on history features
$c_{\theta,i,r}$ and an ego-motion embedding $p_{\theta,i,r}$.
During training, preceding rendered frames supply the history context;
the clean current target is excluded. This teacher-forcing construction
is given in Appendix~\ref{app:video-conditioning}. The video loss is
\begin{equation}
 \Lvideo(i)=\mathbb{E}_{t,\epsilon,r}\!\left[
 \frac{1}{d_z}\left\|D_\phi(x_{i,r,t},t;c_{\theta,i,r},p_{\theta,i,r})
 -\sg(u_{i,r}^\star)\right\|_2^2\right],
 \label{eq:video-fm}
\end{equation}
where $d_z$ is the number of elements in a frame latent and $\sg$
blocks gradients. The expectation averages over frames, flow times,
and noise samples. The loss trains video prediction and the shared
representations used for planning.

\paragraph{Target Selection.}
We first select candidates for rendering by trajectory reward,
then select rendered videos for supervision by prediction error
(Figure~\ref{fig:overview}(c)).
For each valid rendered candidate, we compute the teacher-forced
latent prediction error $E_i^{\rm TF}$ in Eq.~\ref{eq:tf-error},
which combines frame reconstruction and temporal-change errors.
Within each scene, we retain candidates whose error is at or
above the mean over valid rendered candidates.
This subset, denoted Q2, contains high-reward, high-error candidates.
Their matched videos serve as targets for the flow-matching loss
in Eq.~\ref{eq:video-fm}, focusing supervision on visual consequences
that the model predicts less accurately.

\begin{table}[tp]
\centering\small
\caption{\textbf{Planning results on NAVSIM v1 NavTest.}
UniAD results are from \citet{liao2025diffusiondrive}; DriveVA also uses CARLA training data.
Scores for the VLM-based baselines are taken from \citet{zhao2026simwam}.
Epona is adapted to NAVSIM following the official recipe and re-evaluated using the same protocol as \method.}
\label{tab:main-planning}
\setlength{\tabcolsep}{3.5pt}
\begin{adjustbox}{max width=\linewidth}
\begin{tabular}{@{}lrrrrrrrrrr@{}}
\toprule
Method & NC $\uparrow$ & DAC $\uparrow$ & TTC $\uparrow$ & Comfort $\uparrow$ & EP $\uparrow$ & PDMS $\uparrow$\\
\midrule
\rowcolor{tablepurple}
\multicolumn{7}{@{}l}{\textit{Traditional E2E planners}}\\
UniAD \citep{hu2023uniad} & 97.8 & 91.9 & 92.9 & 100.0 & 78.8 & 83.4\\
DiffusionDrive \citep{liao2025diffusiondrive} & 98.2 & 96.2 & 94.7 & 100.0 & 82.2 & 88.1\\
WoTE \citep{li2025wote} & 98.5 & 96.8 & 94.9 & 99.9 & 81.9 & 88.3\\
MeanFuser + BeyondDrive \citep{wang2026beyond} & 98.5 & 98.4 & 95.1 & 100.0 & 84.2 & 90.3\\
ResAD \citep{zheng2026resad} & 98.9 & 97.8 & 94.9 & 100.0 & 87.0 & 90.6\\
\midrule
\rowcolor{tablepurple}
\multicolumn{7}{@{}l}{\textit{VLM-based planners}}\\
AutoVLA \citep{zhou2025autovla} & 98.4 & 95.6 & 98.0 & 99.9 & 81.9 & 89.1\\
Uni-World VLA \citep{liu2026uniworldvla} & 98.7 & 96.7 & 96.1 & 100.0 & 83.2 & 89.4\\
ExploreVLA \citep{sheng2026explorevla} & 98.8 & 98.4 & 96.5 & 99.9 & 83.5 & 90.4\\
ReCogDrive \citep{li2025recogdrive} & 97.9 & 97.3 & 94.9 & 100.0 & 87.3 & 90.8\\
SGDrive \citep{li2026sgdrive} & 98.6 & 97.8 & 96.2 & 100.0 & 85.8 & 91.1\\
\midrule
\rowcolor{tablepurple}
\multicolumn{7}{@{}l}{\textit{World-model-based planners}}\\
LAW \citep{li2025law} & 96.4 & 95.4 & 88.7 & 99.9 & 81.7 & 84.6\\
\epona \citep{zhang2025epona} & 97.8 & 95.0 & 93.9 & 99.9 & 80.3 & 86.2\\
PWM \citep{zhao2025pwm} & 98.6 & 95.9 & 95.4 & 100.0 & 81.8 & 88.1\\
DreamerAD \citep{yang2026dreamerad} & 98.0 & 97.2 & 94.3 & 100.0 & 83.1 & 88.7\\
DriveWAM \citep{shi2026drivewam} & 98.3 & 98.1 & 95.2 & 100.0 & 84.3 & 90.1\\
DriveVA \citep{liu2026driveva} & 99.2 & 97.5 & 98.7 & 100.0 & 83.5 & 90.9\\
\midrule
\rowcolor{blue!12}
\textbf{\method} & 98.0 & 97.2 & 97.0 & 100.0 & 86.7 & \textbf{91.3}\\
\bottomrule
\end{tabular}
\end{adjustbox}
\end{table}
\subsection{Training and Inference}
We use SDE trajectory sampling following Flow-GRPO
\citep{liu2025flowgrpo} and group-relative advantages following GRPO
\citep{shao2024deepseekmath}. The trajectory loss $\Ltraj$ weights clipped transition scores
by these advantages and applies a discount across denoising steps.
Video prediction error does not enter the trajectory advantages.
Our overall training objective combines the trajectory loss,
the video loss on selected targets $\Lvideo^{\rm sel}$,
and the BC regularizer, as defined in Eq.~\ref{eq:total-loss}:
\begin{equation}
 \mathcal{L}=\Ltraj+\lambda_v\Lvideo^{\rm sel}
 +\lambda_{\rm BC}\Lanchor,
 \label{eq:total-loss}
\end{equation}
where $\Lanchor$ is the BC regularizer.
The trajectory loss favors candidates with higher relative
driving rewards, while the BC regularizer trains the model to
follow denoising chains generated by a teacher planner
(Figure~\ref{fig:overview}(b)).
We use the frozen initial trajectory network (TrajDiT) as the
teacher. The teacher and student use the current shared
representation, with gradients stopped on the teacher side.
The weights $\lambda_v$ and $\lambda_{\rm BC}$ control video supervision and
BC regularization. Training repeats sampling, execution, rendering, and model updates. At inference, the model predicts trajectories directly from observation
history, without simulator calls, video generation, or video-based reranking.

\begin{table}[tp]
\centering\small
\caption{\textbf{Planning results on NAVSIM v2 NavTest.}
Epona is re-evaluated using the same protocol as \method.
SGDrive's scores are taken from \citet{zhao2026simwam}. 
}
\label{tab:navsim-v2}
\setlength{\tabcolsep}{2.5pt}
\begin{adjustbox}{max width=\linewidth}
\begin{tabular}{@{}lcccccccccc@{}}
\toprule
Method & NC $\uparrow$ & DAC $\uparrow$ & DDC $\uparrow$ & TLC $\uparrow$ & EP $\uparrow$ & TTC $\uparrow$ & LK $\uparrow$ & HC $\uparrow$ & EC $\uparrow$ & EPDMS $\uparrow$\\
\midrule
\rowcolor{tablepurple}
\multicolumn{11}{@{}l}{\textit{Traditional E2E planners}}\\
ResAD \citep{zheng2026resad} & 97.8 & 97.2 & 99.5 & 99.8 & 88.2 & 96.9 & 97.0 & 98.4 & 88.2 & 85.5\\
DriveSuprim \citep{yao2026drivesuprim} & 98.4 & 98.6 & 99.6 & 99.8 & 90.5 & 97.8 & 97.0 & 98.3 & 78.6 & 87.1\\
DiffRefiner \citep{yin2026diffrefiner} & 98.6 & 98.5 & 99.7 & 99.9 & 87.6 & 98.1 & 98.1 & 98.3 & 87.9 & 87.4\\
\midrule
\rowcolor{tablepurple}
\multicolumn{11}{@{}l}{\textit{VLM-based planners}}\\
SGDrive \citep{li2026sgdrive} & 98.6 & 94.3 & 99.5 & 99.9 & 86.0 & 97.9 & 96.1 & 98.3 & 85.9 & 86.2\\
DriveWorld-VLA \citep{jia2026driveworldvla} & 98.6 & 99.1 & 99.6 & 99.8 & 87.4 & 97.9 & 97.0 & 97.8 & 78.6 & 86.8\\
ELF-VLA-8B \citep{luo2026elfvla} & 98.9 & 98.1 & 99.4 & 99.8 & 88.5 & 98.4 & 96.9 & 98.3 & 87.2 & 87.1\\
\midrule
\rowcolor{tablepurple}
\multicolumn{11}{@{}l}{\textit{World-model-based planners}}\\
\epona \citep{zhang2025epona} & 97.2 & 95.6 & 99.3 & 99.7 & 88.4 & 96.4 & 96.9 & 97.9 & 67.6 & 85.0\\
SafeDrive \citep{kim2026safedrive} & 99.5 & 99.0 & 99.7 & 99.9 & 88.6 & 98.9 & 97.5 & 98.2 & 81.9 & 87.5\\
DreamerAD \citep{yang2026dreamerad} & 98.0 & 97.2 & 99.5 & 99.8 & 87.8 & 97.4 & 97.5 & 98.3 & 72.4 & 87.7\\
\midrule
\rowcolor{blue!12}
\textbf{\method} & 98.5 & 98.5 & 99.8 & 99.0 & 90.8 & 98.5 & 97.5 & 99.0 & 60.0 & \textbf{88.2}\\
\bottomrule
\end{tabular}
\end{adjustbox}
\end{table}
\section{Experiments}
\label{sec:experiments}
\subsection{Experimental Setup}
\paragraph{Implementation Details.}
We initialize \method from an Epona checkpoint adapted to NAVSIM as
described in Appendix~\ref{app:algorithm}, and perform post-training
on NavTrain samples with available World Engine scene reconstructions. We use the 3DGS-based simulator in
World Engine \citep{li2026worldengine} to render matched future videos.
In \method (Exp2), the video flow-matching loss uses Q2 targets and
updates both the video network (VideoDiT) and the shared encoder.
All ablations retain trajectory GRPO and the BC regularizer, with
trajectory GRPO updating the shared encoder throughout. The mixed
high/low-reward variants, Exp4 and Exp6, additionally apply the
safety-specific advantage treatment described in
Appendix~\ref{app:controls}.
\paragraph{Benchmarks and Metrics.}
We evaluate \method on the NavTest splits of NAVSIM v1 and v2
\citep{Dauner2024navsim,Cao2025navsimv2}, using PDMS and
EPDMS, respectively, to assess safety, progress, and comfort.
On HUGSIM \citep{zhou2024hugsim}, we report route completion (RC)
and HD-Score, which combines RC with safety and comfort.
Training uses the reward in Eq.~\ref{eq:training-reward};
Benchmark scores follow their respective evaluation protocols.
We report results averaged over three independent runs with different random seeds.
Metric details are provided in Appendix~\ref{app:benchmarks}.

\begin{table}[!htbp]
\centering\small
\caption{\textbf{Planning results on HUGSIM.}
E/M/H/X denote easy, medium, hard, and extreme scenarios.
Baseline results are taken from Driving on Registers, Table 2.
Higher values are better for all metrics.}
\label{tab:hugsim}
\begin{adjustbox}{max width=\linewidth}
\begin{tabular}{@{}lrrrrrrrrrr@{}}
\toprule
& \multicolumn{5}{c}{RC $\uparrow$}
& \multicolumn{5}{c}{HD-Score $\uparrow$} \\
\cmidrule(lr){2-6}\cmidrule(lr){7-11}
Method & E & M & H & X & Avg.
       & E & M & H & X & Avg. \\
\midrule
VAD & 51.3 & 31.1 & 25.3 & 26.5 & 31.4
    & 36.3 & 9.5 & 8.0 & 11.5 & 13.4 \\
LTF & 67.8 & 35.1 & 26.2 & 40.5 & 38.9
    & 58.9 & 18.0 & 9.8 & 25.9 & 23.7 \\
UniAD & 78.4 & 60.5 & 33.6 & 17.8 & 45.9
      & 64.9 & 45.8 & 20.6 & 6.6 & 32.7 \\
DrivoR & 80.9 & 50.5 & 33.8 & 47.1 & 49.8
       & 73.3 & 34.6 & 18.8 & 32.5 & 35.7 \\
\midrule
\rowcolor{blue!12}
\textbf{\method}
& 78.8 & 57.6 & 38.3 & 40.1 & \textbf{51.5}
& 69.9 & 45.0 & 23.3 & 27.2 & \textbf{38.8} \\
\bottomrule
\end{tabular}
\end{adjustbox}
\end{table}

\subsection{Main Results}
\paragraph{NAVSIM.}
Starting from the NAVSIM-adapted Epona initialization at 86.2 PDMS,
\method reaches 91.3 (+5.1) on NAVSIM v1
(Figure~\ref{fig:planning-gains}a; Table~\ref{tab:main-planning}).
Ego progress rises from 80.3 to 86.7, time-to-collision compliance
from 93.9 to 97.0, and drivable-area compliance from 95.0 to 97.2.
On NAVSIM v2, \method improves EPDMS from 85.0 to 88.2 (+3.2),
and achieves 90.8 ego progress and 98.5 for both
drivable-area and time-to-collision compliance
(Table~\ref{tab:navsim-v2}).

\begin{figure}[t]
\centering
\includegraphics[width=\linewidth]{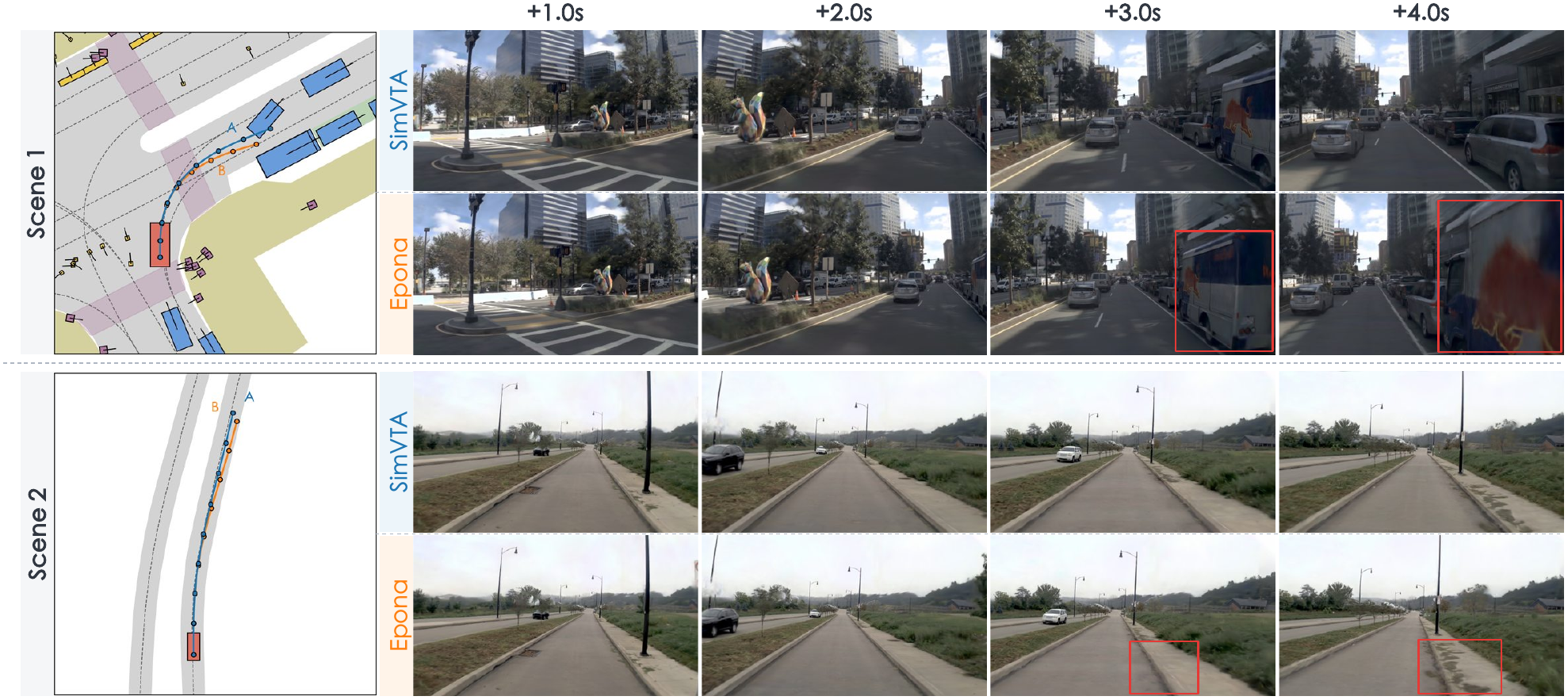}
\caption{\textbf{Qualitative planning comparison.}
Left: planned trajectories of \method (A, blue) and Epona (B, orange).
Right: World Engine renderings along the corresponding paths at 1--4\,s.
Red boxes highlight Epona's collision and road-departure regions.}
\label{fig:planning-cases}
\end{figure}

\method also improves planning in scenes where the Epona initialization
fails (Figure~\ref{fig:planning-cases}). In these examples, Epona's
trajectories lead to a collision or leave the drivable area, whereas
\method avoids the collision and stays within the drivable area.

\paragraph{HUGSIM.}
We further evaluate the final planner in closed-loop driving on the
same 345 pre-challenge HUGSIM scenarios, using the same corrected
evaluator and controller as \citet{kirby2026drivor}.
\method achieves the highest average RC (51.5) and HD-Score (38.8)
among the compared methods, exceeding the reported DrivoR results
by 1.7 and 3.1 points, respectively (Table~\ref{tab:hugsim}).
It also leads on both metrics in hard scenarios, reaching 38.3 RC
and 23.3 HD-Score.

\begin{figure}[!htbp]
\centering
\includegraphics[width=.965\linewidth]{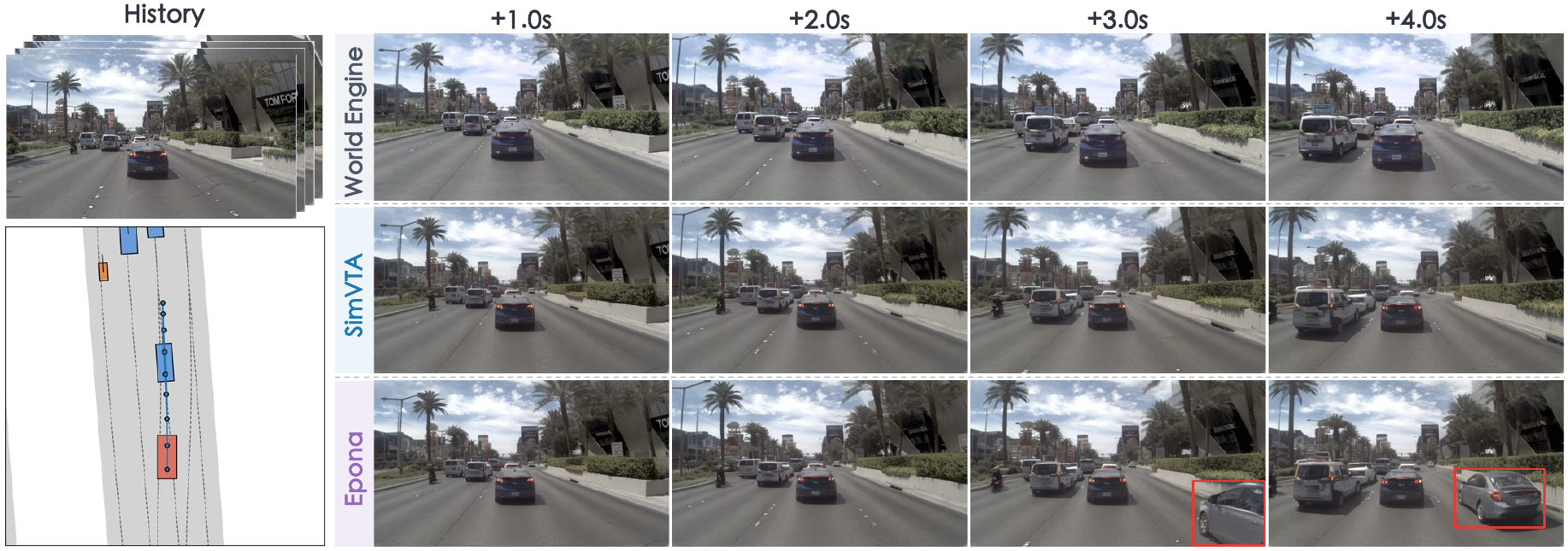}
\caption{\textbf{Video--trajectory alignment before and after post-training.}
Left: observation history and trajectory in BEV.
Right: World Engine renderings and predictions from
\method and Epona.}
\label{fig:video-trajectory-alignment}
\end{figure}

\paragraph{Video-Trajectory Alignment.}
Beyond the planning gains, we compare \method with Epona to examine
whether post-training improves the consistency of predicted videos
with the scene along a trajectory
(Figure~\ref{fig:video-trajectory-alignment}).
Epona generates an additional vehicle on the right
(highlighted by red boxes) that is absent from both the bird's-eye view (BEV) and
the corresponding World Engine renderings.
\method avoids this inconsistency in the illustrated scene.

\begin{wraptable}{r}{0.52\textwidth}
    \centering\small
    \setlength{\tabcolsep}{3pt}
    \setlength{\abovecaptionskip}{0pt}
    \caption{\textbf{Video quality.}}
    \label{tab:video-quality}
    \begin{tabularx}{\linewidth}{@{}>{\raggedright\arraybackslash}Xrr@{}}
    \toprule
    Source & FID $\downarrow$ & FVD $\downarrow$\\
    \midrule
    \rowcolor{tablepurple}
    \multicolumn{3}{@{}l@{}}{\textit{3DGS renderings}}\\
    World Engine & 7.6 & 21.8\\
    \midrule
    \rowcolor{tablepurple}
    \multicolumn{3}{@{}l@{}}{\textit{WAM predictions}}\\
    Trajectory GRPO only (Exp0) & 20.4 & 105.4\\
    \textbf{SimVTA} (Exp2) & 13.1 & 46.3\\
    \bottomrule
    \end{tabularx}
\end{wraptable}

\paragraph{Video Quality.}
We evaluate both World Engine renderings and autoregressive model predictions
against real videos on NAVSIM v1 (Table~\ref{tab:video-quality}).
Lower Fréchet Inception Distance (FID) and Fréchet Video Distance (FVD)
indicate better video quality.
Along logged GT trajectories, World Engine achieves an FID of 7.6 and an
FVD of 21.8.
Compared with trajectory GRPO alone (Exp0), \method reduces FID
from 20.4 to 13.1 and FVD from 105.4 to 46.3, improving video
prediction quality.

\begin{figure}[!htbp]
\centering
\includegraphics[width=.965\linewidth]{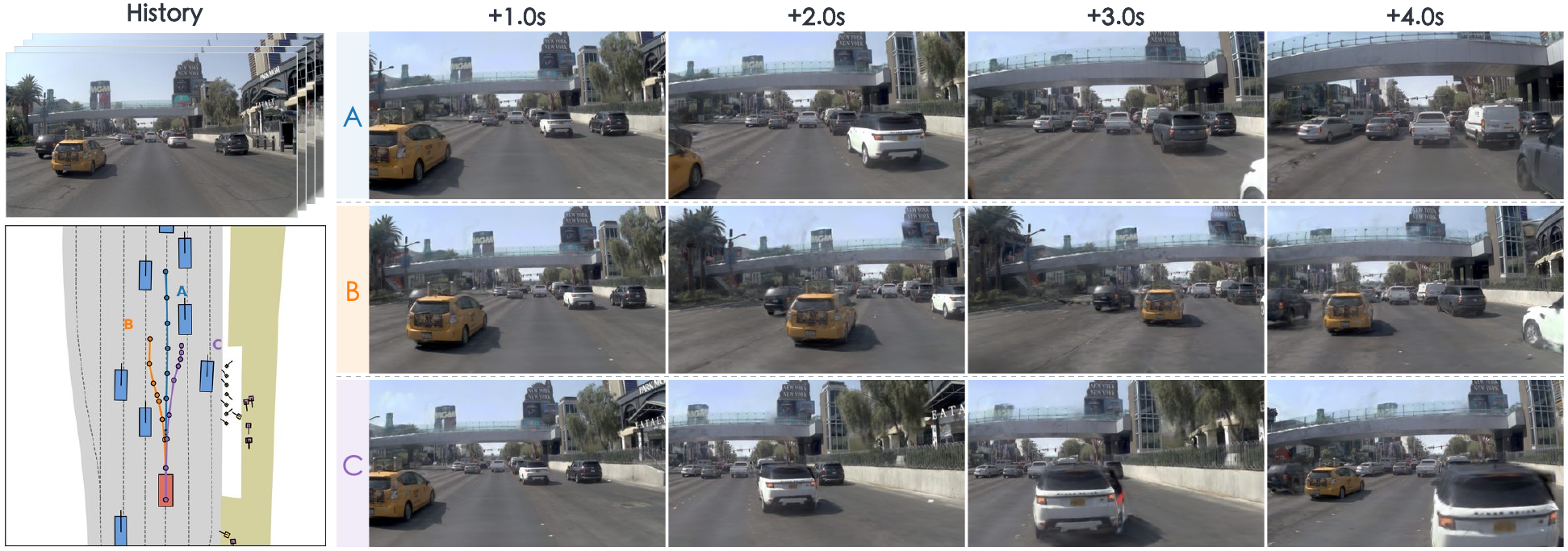}
\caption{\textbf{Matched futures for alternative actions.}
Left: shared observation history and executed paths for going straight (A),
changing lanes to the left (B), and changing lanes to the right (C).
Right: corresponding future frames rendered by the simulator.
Colors match the paths and image rows.}
\label{fig:matched-renderings}
\end{figure}
\subsection{Ablation Studies}
\paragraph{Updating the Shared Representations.}
Video supervision improves planning when its gradients reach the
shared encoder. Restricting video gradients to VideoDiT (Exp1)
yields 89.1 PDMS, matching trajectory GRPO alone (Exp0;
Figure~\ref{fig:planning-gains}a). Allowing video gradients to also
update the shared encoder (Exp2) raises PDMS to 91.3 (+2.2)
(Table~\ref{tab:ablation-shared-encoder}).

\paragraph{Selecting Informative Futures.}
We compare video supervision across four candidate groups:
Q1 (high reward, low error), Q2 (high reward, high error),
Q3 (low reward, low error), and Q4 (low reward, high error).
High- and low-reward candidates come from the top and bottom
of the trajectory reward ranking, respectively.
Within each scene, errors below the mean over valid rendered
candidates are low; those at or above it are high.

\textbf{Q2 supervision yields the highest PDMS among the four settings.}
Exp2 uses Q2 supervision and achieves 91.3 PDMS.
Exp5 uses the same high-reward rendering pool but expands
video supervision to Q1+Q2, yielding 91.0 PDMS.
Under the same high-reward rendering pool, focusing supervision on
the higher-error Q2 subset performs better than supervising Q1+Q2.
Exp4 and Exp6 use mixed high/low-reward pools with Q2+Q3
and Q1--Q4 supervision, respectively, both achieving 90.9 PDMS
(Table~\ref{tab:ablation-target-selection}).
These two variants also apply the safety-specific advantage
treatment described in Appendix~\ref{app:controls}.

\begin{table}[htbp]
\centering
\begin{minipage}[t]{0.48\linewidth}
\vspace{0pt}
\centering\small
\setlength{\tabcolsep}{1pt}
\caption{\textbf{Ablation on shared encoder updates.}}
\label{tab:ablation-shared-encoder}
\begin{tabularx}{\linewidth}{@{}l>{\raggedright\arraybackslash}Xr>{\columncolor{tableblue}}r@{}}
\toprule
Setting & Video loss updates & NC $\uparrow$ & PDMS $\uparrow$\\
\midrule
Exp0 & None & 97.1 & 89.1\\
Exp1 & VideoDiT & 97.6 & 89.1\\
\textbf{Exp2} & VideoDiT + shared encoder & 98.0 & 91.3\\
\bottomrule
\end{tabularx}
\end{minipage}\hfill
\begin{minipage}[t]{0.48\linewidth}
\vspace{0pt}
\centering\small
\setlength{\tabcolsep}{3pt}
\caption{\textbf{Ablation on video target selection.}}
\label{tab:ablation-target-selection}
\begin{tabularx}{\linewidth}{@{}l>{\raggedright\arraybackslash}Xr>{\columncolor{tableblue}}r@{}}
\toprule
Setting & Targets & NC $\uparrow$ & PDMS $\uparrow$\\
\midrule
Exp4 & Q2 + Q3 & 97.4 & 90.9\\
Exp6 & Q1 + Q2 + Q3 + Q4  & 97.7 & 90.9\\
Exp5 & Q1 + Q2 & 98.0 & 91.0\\
\textbf{Exp2} & Q2 & 98.0 & 91.3\\
\bottomrule
\end{tabularx}
\end{minipage}
\end{table}

\begin{table}[H]
\centering
\begin{minipage}[t]{0.48\linewidth}
\vspace{0pt}
\centering\small
\setlength{\tabcolsep}{3pt}
\caption{\textbf{Video-training variants.}\strut}
\label{tab:ablation-video-objective}
\begin{tabularx}{\linewidth}{@{}l>{\raggedright\arraybackslash}Xr>{\columncolor{tableblue}}r@{}}
\toprule
Setting & Video supervision & NC $\uparrow$ & PDMS $\uparrow$\\
\midrule
Exp8 & All valid rendered + video GRPO & 97.9 & 91.0\\
\textbf{Exp2} & Q2 + flow matching & 98.0 & 91.3\\
\bottomrule
\end{tabularx}
\end{minipage}\hfill
\begin{minipage}[t]{0.48\linewidth}
\vspace{0pt}
\centering\small
\setlength{\tabcolsep}{3pt}
\caption{\textbf{Online versus offline rendering.}}
\label{tab:online-offline}
\begin{tabularx}{\linewidth}{@{}l>{\raggedright\arraybackslash}X>{\columncolor{tableblue}}r@{}}
\toprule
Setting & Rendering strategy & PDMS $\uparrow$\\
\midrule
Exp0 & No video supervision & 89.1\\
Exp9 & Fixed offline cache & 90.4\\
\textbf{Exp2} & Online rendering & 91.3\\
\bottomrule
\end{tabularx}
\end{minipage}
\end{table}

\paragraph{Comparing Video-Training Variants.}
We further compare our selective video supervision with directly
applying video GRPO to the rendered candidates.
Under the same rendering budget of four high-reward candidates,
Exp8 applies video GRPO to all valid rendered candidates and obtains
91.0 PDMS. In contrast, Exp2 first selects the higher-error Q2 subset
according to $E_i^{\rm TF}$ and then optimizes the selected targets
with the flow-matching loss, reaching 91.3 PDMS
(Table~\ref{tab:ablation-video-objective}).
Both variants allow video gradients to update the shared encoder.
This comparison tests whether selecting informative video targets
before supervision is more effective than directly applying video GRPO
to all valid rendered candidates.

\paragraph{Comparing Online and Offline Rendering.}
We compare online rendering with a fixed offline cache
(Table~\ref{tab:online-offline}). Exp9 follows the same settings as Exp2, except that
trajectories are sampled offline from the initial policy and their
rendered videos are cached and reused throughout post-training.
Exp2 instead refreshes video targets for current-policy trajectories.
Exp9 achieves 90.4 PDMS, compared with 89.1 for trajectory-only Exp0
and 91.3 for Exp2, showing a 0.9-point advantage for refreshing video
targets from the current policy over using a fixed initial-policy cache.

\paragraph{Preserving Trajectory-Video Correspondence.}
To test whether the benefit depends on correct trajectory-video pairing
rather than video supervision alone, Exp3 shuffles rendered targets
across trajectory conditions within the same scene after selection.
Breaking this alignment reduces PDMS from 91.3 (Exp2) to 83.1
(Exp3; Table~\ref{tab:ablation-correspondence}), demonstrating the
importance of preserving trajectory-video correspondence.
Figure~\ref{fig:matched-renderings} shows the matched setting, where
each target video is rendered along its corresponding path.

\paragraph{Increasing the Rendering Budget.}
Doubling the rendering budget provides no PDMS gain:
Exp7 achieves 91.2 versus 91.3 for Exp2 with the same target selection
(Figure~\ref{fig:planning-gains}c; Table~\ref{tab:ablation-rendering-budget}).
\begin{table}[htbp]
\centering
\begin{minipage}[t]{0.48\linewidth}
\vspace{0pt}
\centering\small
\setlength{\tabcolsep}{3pt}
\caption{\textbf{Trajectory--video correspondence.}\strut}
\label{tab:ablation-correspondence}
\begin{tabularx}{\linewidth}{@{}l>{\raggedright\arraybackslash}Xr>{\columncolor{tableblue}}r@{}}
\toprule
Setting & Video targets & NC $\uparrow$ & PDMS $\uparrow$\\
\midrule
Exp3 & Shuffled & 94.1 & 83.1\\
\textbf{Exp2} & Matched & 98.0 & 91.3\\
\bottomrule
\end{tabularx}
\end{minipage}\hfill
\begin{minipage}[t]{0.48\linewidth}
    \vspace{0pt}
    \centering\small
    \setlength{\tabcolsep}{3pt}
    \caption{\textbf{Rendering budget.}}
    \label{tab:ablation-rendering-budget}
    \begin{tabularx}{\linewidth}{@{}lXrr>{\columncolor{tableblue}}r@{}}
    \toprule
    Setting & Targets & Budget & NC $\uparrow$ & PDMS $\uparrow$\\
    \midrule
    Exp7 & Q2 & $2\times$ & 97.8 & 91.2\\
    \textbf{Exp2} & Q2 & $1\times$ & 98.0 & 91.3\\
    \bottomrule
    \end{tabularx}
    \end{minipage}
\end{table}

\section{Conclusion}
We presented \method, a simulation-grounded post-training framework
that combines trajectory GRPO with video supervision.
Rendering futures along current-policy trajectories provides visual
targets beyond logged action-future pairs and jointly trains shared
representations through video prediction and trajectory optimization.
The full framework improves NAVSIM v1 PDMS from 86.2
to 91.3 (+5.1) and achieves 88.2 EPDMS on NAVSIM v2 and 38.8 average HD-Score
on HUGSIM. Ablations show that updating shared representations with video
supervision improves planning, while prioritizing higher-error futures
among high-reward trajectories further improves performance.

Limitations and directions for future work are discussed in
Appendix~\ref{app:limitations}.

\section*{Acknowledgments}
We would like to thank other members of OpenDriveLab for their
valuable discussions and technical support.

\bibliography{short_bib,main}
\bibliographystyle{preprint}
\clearpage
\noindent\textbf{\textit{\LARGE Appendix}}
\vskip8pt
\appendix
\section{Training Procedure}
\label{app:algorithm}

Before applying \method, we fine-tune the nuPlan-pretrained Epona
checkpoint on NAVSIM following the official adaptation recipe to
adapt it from 10 Hz to 2 Hz. The adapted checkpoint initializes
both Exp0 and Exp2. We evaluate it using the same NAVSIM v1 protocol
as the post-trained models and obtain 86.2 PDMS, matching the overall
score reported for Epona. Unless stated otherwise, the implementation
details below describe Exp2.

\paragraph{Training Loop.}
\begin{enumerate}
\item Encode observation histories and sample $G$ trajectories,
retaining their normalized denoising states.
\item Execute each candidate with PDM, compute its training reward,
and normalize rewards within each observation group.
\item Check execution validity, pose support, and safety. Rank eligible
candidates by reward and select the rendering pool.
\item Render the selected paths and encode their future frames,
retaining the candidate-to-video correspondence.
\item Compute detached teacher-forced errors and classify the
valid rendered candidates into quadrants.
\item Select the video-training subset and future frame for this update.
Generate the detached BC teacher chain from the current shared representation.
\item Combine the trajectory loss, selected video loss, and BC regularizer.
Update the trainable parameters and repeat with the updated policy.
\end{enumerate}

\paragraph{Architecture and Temporal Settings.}
Epona's spatiotemporal Transformer is the shared encoder, paired with
TrajDiT and VideoDiT. Four observed frames condition eight trajectory
waypoints spanning four seconds. PDM evaluates the path at 10 Hz;
video pose conditions are sampled at 2 Hz. Exp2 computes teacher‑forced prediction errors over future frames within the planning window and trains one sampled future frame per update.

\paragraph{Trajectory Reward.}
Each candidate is scored together with the cached PDM reference trajectory.
The training reward is
\begin{equation}
 R_i=\mathrm{NC}_i\,\mathrm{DAC}_i
 \frac{10\,\mathrm{EP}_i+5\,\mathrm{TTC}_i+
 2\,\mathrm{Comfort}_i}{17}.
 \label{eq:training-reward}
\end{equation}
Following ReCogDrive \citep{li2025recogdrive}, we use weights
of 10, 5, and 2 for EP, TTC, and Comfort, respectively,
with the higher EP weight encouraging efficient driving.
The component scores lie in $[0,1]$. Progress follows the PDM scorer's
normalization over the candidate and reference trajectories.
Driving-direction compliance is used for candidate screening but has
zero weight in this reward. Exp2 excludes video prediction error from trajectory rewards.

\paragraph{Candidate Validity and Safety.}
An eligible video candidate must execute successfully and have finite
pose changes within VideoDiT's support at every 2 Hz step:
$0\leq\Delta x\leq16$ m, $|\Delta y|\leq1.5$ m, and
$|\Delta\mathrm{yaw}|\leq32^\circ$.
Its maximum reverse speed must not exceed $10^{-3}$ m/s.
For Exp2, the NC, DAC, TTC, and DDC scores
must each be at least 0.999. We rank eligible candidates by $R_i$,
break ties by sampling index, and select the first $K$.
If fewer than $K$ qualify, only those candidates are rendered.
Failed renders are excluded from video supervision.
A scene skips video supervision if all candidates fail the safety
checks or its maximum reward is at most $10^{-6}$.
These filters do not remove candidates from the Exp2 trajectory loss.

\paragraph{Optimization Configuration.}
Table~\ref{tab:training-config} lists the video-supervision settings
for Exp2 and the optimization defaults.
Training uses AdamW, implemented by DeepSpeedCPUAdam in AdamW mode
when optimizer offloading is enabled. Bias and normalization parameters
are excluded from weight decay. The learning rate is halved at steps
300 and 1500. The shared encoder uses the same learning rate as the heads.
When no video target participates across all devices, VideoDiT
parameters remain unchanged for that update.

\begin{table}[t]
\centering\small
\caption{Exp2 video-supervision settings and default optimization parameters.}
\label{tab:training-config}
\begin{tabular}{@{}ll@{}}
\toprule
Parameter & Value\\
\midrule
Sampled trajectories / rendering budget & $G=32$ / $K=4$\\
Batch size per GPU / default random seed & 4 / 1234\\
Training hardware & 16 NVIDIA H20 GPUs\\
Initial learning rate / AdamW betas & $10^{-5}$ / $(0.9,0.95)$\\
Weight decay / LR decay factor & 0.01 / 0.5\\
LR decay steps & 300, 1500\\
Trajectory denoising steps / noise scale & 10 / 0.3\\
Diffusion-coefficient floor / denoising discount & 0.01 / 0.6\\
Video loss / BC weights & $\lambda_v=0.1$ / $\lambda_{\rm BC}=0.1$\\
Video gradient scale into shared parameters & $\eta=0.2$\\
Video-error horizon / temporal discount & $H=8$ / $\gamma_{\rm v}=0.85$\\
Video-training candidates per scene per update & All valid Q2\\
Trained future frames per candidate per update & 1\\
Mixed precision / distributed optimizer & bfloat16 / ZeRO-2, CPU offload\\
\bottomrule
\end{tabular}
\end{table}

\section{Conditional Video Prediction}
\label{app:video-conditioning}
The next-frame prediction head VideoDiT is named VisDiT in Epona.
The process $W_\phi$ applies the velocity network $D_\phi$ and decodes
the resulting image latents. Flow time $t$ in Eq.~\ref{eq:flow-path}
runs from noise at zero to the target at one and is distinct from frame
index $r$. The factor $1/d_z$ in Eq.~\ref{eq:video-fm} averages the
squared error over the elements of one frame latent.

\paragraph{Internal Motion Conditioning.}
For each selected candidate, World Engine renders the supervision
target along its executed path. The WAM uses the same executed pose
sequence for its internal video conditioning:
\begin{equation*}
 V_i^{\rm sim}=\mathcal{S}(o,\tilde\tau_i),\qquad
 V_i^{\rm pred}=W_\phi(h_\theta(o),\tilde\tau_i).
\end{equation*}
This pairing is used during post-training; the motion conditioning
is internal to the WAM.

To predict frame $r$, we construct conditions from a rolling history
window containing observed or preceding rendered frames and pose changes:
\begin{equation}
 c_{\theta,i,r}=h_\theta(o,z^{\rm sim}_{i,<r},\sg(\Delta\tilde\tau_{i,<r})),
 \qquad p_{\theta,i,r}=e_\theta(\sg(\Delta\tilde\tau_{i,r})).
 \label{eq:teacher-condition}
\end{equation}
Here $<r$ denotes preceding frames, $\Delta\tilde\tau_{i,r}$ is the
executed pose change, and $e_\theta$ embeds that motion.
The expression for $h_\theta$ denotes the rolling context rather than
an ever-growing sequence. The clean current target is excluded from
this context. Rendered latents and pose inputs are fixed; the shared
encoder and motion embedding remain trainable.

\paragraph{Teacher-forced Prediction Error.}
We compute the prediction error at $t_\star=0.5$ using one Gaussian tensor
$\epsilon_o$ shared by all candidates and future frames of an observation:
\begin{equation}
 \begin{aligned}
 x^\star_{i,r}&=t_\star z^{\rm sim}_{i,r}+(1-t_\star)\epsilon_o,\\
 \hat z_{i,r}&=x^\star_{i,r}+(1-t_\star)
 D_\phi(x^\star_{i,r},t_\star;c_{\theta,i,r},p_{\theta,i,r}),\\
 e_{i,r}&=\hat z_{i,r}-z^{\rm sim}_{i,r}.
 \end{aligned}
\end{equation}
With $\alpha_r=\gamma_{\rm v}^{r-1}/\sum_{j=1}^{H}\gamma_{\rm v}^{j-1}$,
we define
\begin{equation}
 \begin{aligned}
 E_i^{\rm TF}
 &=0.7\sum_{r=1}^{H}\alpha_r\frac{\|e_{i,r}\|_2^2}{d_z}\\
 &\quad+0.3\sum_{r=2}^{H}
 \frac{\alpha_r}{\sum_{j=2}^{H}\alpha_j}
 \frac{\|\sqrt{\omega_{i,r}}\odot(e_{i,r}-e_{i,r-1})\|_2^2}{d_z}.
 \end{aligned}
 \label{eq:tf-error}
\end{equation}
For token $p$ and channel $c$, the motion weight is
\begin{equation}
 \begin{aligned}
 m_{i,r,p}&=\operatorname{mean}_{c}
 |z^{\rm sim}_{i,r,p,c}-z^{\rm sim}_{i,r-1,p,c}|,\\
 \omega_{i,r,p}&=\operatorname{clip}\left(
 \frac{m_{i,r,p}}{\max(\operatorname{mean}_{p'}m_{i,r,p'},10^{-6})},
 0.25,4\right).
 \end{aligned}
\end{equation}
The weight is broadcast over channels. We use hyperparameters $H$ and $\gamma_{\rm v}=0.85$. The complete error computation is detached.
Unlike the video‑training loss, this score uses a fixed flow time,
shared evaluation noise, and a temporal‑change term.

\section{Video Supervision and Policy Optimization}
\label{app:objectives}
\paragraph{Quadrant Assignment and Training Subset.}
Reward labels follow the ranked rendering pool: Exp2 uses the
high-reward pool, while mixed-pool variants retain separate high-
and low-reward labels. No reward-mean threshold is applied.
The error threshold is the mean of $E_i^{\rm TF}$ over valid
rendered candidates within each scene. An error below the mean is
low; an error equal to or above it is high.

For each scene, Exp2 trains on all valid Q2 candidates at every update.
At training step $s$, the trained future frame for each candidate is
$1+(s\bmod H)$.
Preceding target frames still provide context when only a later frame
is trained. Flow time and training noise are sampled independently
for each selected latent. Candidate selection is detached.

\paragraph{Video-gradient Masks.}
Let $m_i^{\rm head}$ include candidate $i$ in video training, and let
$m_i^{\rm shared}\leq m_i^{\rm head}$ additionally allow gradients
into the shared parameters. We use
$g_\eta(x)=\sg(x)+\eta(x-\sg(x))$, which preserves the forward value
and scales its backward gradient by $\eta$.
For the current update, we estimate Eq.~\ref{eq:video-fm} at the
cyclic frame $r=1+(s\bmod H)$ using newly sampled flow time and noise.
Write $\Lvideo^{\rm joint}(i)$ for this estimate with
$g_\eta(c_{\theta,i,r})$ and $g_\eta(p_{\theta,i,r})$ as conditions,
and $\Lvideo^{\rm head}(i)$ for the estimate with both conditions detached.
Then
\begin{equation}
 \Lvideo^{\rm sel}=\frac{1}{Z}\sum_i\left[
 m_i^{\rm shared}\Lvideo^{\rm joint}(i)
 +(m_i^{\rm head}-m_i^{\rm shared})\Lvideo^{\rm head}(i)\right],
 \label{eq:masked-video}
\end{equation}
where $Z=\max(1,\sum_i m_i^{\rm head})$ and the sum covers
participating candidates across all devices.
In Exp2, both masks select Q2 candidates and $\eta=0.2$.
This scale affects only video gradients entering the shared encoder
and motion embedding; trajectory, BC, and VideoDiT gradients retain
their own loss weights. The video loss is zero when no target participates.

\paragraph{Trajectory GRPO.}
We use group-relative advantages with a clipped transition-score objective.
Let $a_i^{(k)}$ be the normalized trajectory state at step $k$,
with $N$ transitions and $d_\tau$ scalar coordinates.
For a stored transition $(a,a')$, the sampler specifies a mean
$\mu_{\theta,\psi,k}(a,o)$ and standard deviation $\sigma_k$.
The implemented score is
\begin{equation}
 \ell_{\theta,\psi,k}(a,a';o)=\frac{1}{d_\tau}
 \sum_{d=1}^{d_\tau}\operatorname{clip}\left(
 -\frac{(a'_d-\mu_{\theta,\psi,k,d}(a,o))^2}{2\sigma_k^2},-5,2\right).
 \label{eq:transition-score}
\end{equation}
This score omits the Gaussian normalizing term and is clipped before
averaging over coordinates. For each observation, we compute
\begin{equation}
 \begin{aligned}
 \bar R&=\frac{1}{G}\sum_{j=1}^{G}R_j,\\
 s_R^2&=\frac{1}{G-1}\sum_{j=1}^{G}(R_j-\bar R)^2,\\
 A_i&=\frac{R_i-\bar R}{s_R+\varepsilon_{\rm adv}},\\
 \Ltraj&=-\mathbb{E}_o\left[
 \frac{1}{GN}\sum_{i=1}^{G}\sum_{k=0}^{N-1}
 \gamma_{\rm d}^{N-1-k}\sg(A_i)
 \ell_{\theta,\psi,k}(a_i^{(k)},a_i^{(k+1)};o)\right].
 \end{aligned}
 \label{eq:trajectory-loss}
\end{equation}
We use $\varepsilon_{\rm adv}=10^{-8}$ and $\gamma_{\rm d}=0.6$.
Sampled states and advantages are detached; gradients through the
transition mean update the shared encoder and TrajDiT.
The trajectory objective uses no current-to-sampling-policy probability ratio.

\paragraph{SDE Transitions.}
Let $s_0=1$ and $s_N=0$, with a decreasing grid between them.
For transitions $k=0,\ldots,N-1$, define
$\Delta s_k=s_{k+1}-s_k<0$. The trajectory velocity network
$v_\psi$ uses ascending flow time $1-s_k$. Define
\begin{equation}
 \begin{aligned}
 \bar s_k&=\begin{cases}s_1,&k=0,\\s_k,&k>0,\end{cases}\\
 \beta_k&=\max\left(\beta_{\min},
 \nu\sqrt{\frac{s_k}{1-\bar s_k}}\right),\\
 \mu_{\theta,\psi,k}(a,o)
 &=a\left(1+\frac{\beta_k^2}{2s_k}\Delta s_k\right)\\
 &\quad-v_\psi(a,1-s_k;h_\theta(o))
 \left(1+\frac{\beta_k^2(1-s_k)}{2s_k}\right)\Delta s_k,\\
 \sigma_k&=\beta_k\sqrt{-\Delta s_k}.
 \end{aligned}
 \label{eq:trajectory-sde}
\end{equation}
Starting from Gaussian noise, we sample
$a^{(k+1)}=\mu_{\theta,\psi,k}+\sigma_k\operatorname{clip}(\xi_k,-5,5)$
with independent $\xi_k\sim\mathcal{N}(0,I)$.
We use $\nu=0.3$, $\beta_{\min}=0.01$, and $N=10$.
The shifted time grid is
$s_k=e^m u_k/(e^m u_k+1-u_k)$, where $u_k=1-k/N$ and
$m=0.5+(1.15-0.5)(8-256)/(4096-256)$.
Before denormalization and reward evaluation, final longitudinal,
lateral, and yaw action coordinates are clipped to
$[-1.7,1.7]$, $[-2.6,2.6]$, and $[-10,10]$, respectively.
These bounds are in normalized coordinates. The inverse mapping is
$x=25(a_x+1)$ m, $y=10a_y$ m, and
$\mathrm{yaw}=12a_{\mathrm{yaw}}$ degrees, with waypoints expressed
relative to the current ego frame.
Stored policy transitions retain the unclipped final state.

\paragraph{BC Regularizer.}
Let $\psi_0$ denote the frozen initial TrajDiT parameters.
For each observation, one Gaussian initial state generates an ODE chain
$b^{(0)},\ldots,b^{(N)}$ using $\psi_0$ and $\sg(h_\theta(o))$.
The teacher follows
$b^{(k+1)}=b^{(k)}-\Delta s_k
v_{\psi_0}(b^{(k)},1-s_k;\sg(h_\theta(o)))$.
Its final state uses the action-clipping bounds above. We optimize
\begin{equation}
 \Lanchor=-\mathbb{E}_{o,b^{(0)}}\left[
 \frac{1}{N}\sum_{k=0}^{N-1}
 \ell_{\theta,\psi,k}(\sg(b^{(k)}),\sg(b^{(k+1)});o)\right].
 \label{eq:bc-regularizer}
\end{equation}
The teacher parameters remain fixed, while teacher chains are regenerated
from the current shared representation. Student-side gradients update
the shared encoder and TrajDiT. The BC objective averages over steps
without the trajectory loss's denoising discount.

\section{Experimental Variants and Budget Accounting}
\label{app:controls}
All ablations use Exp2 as the reference setting.
Paper experiment numbers differ from the implementation entry names:
Exp5, Exp6, and Exp8 correspond to \texttt{exp8\_high4\_all},
\texttt{exp11\_high2\_low2\_all}, and
\texttt{exp10\_high4\_video\_grpo}, respectively.
Table~\ref{tab:video-budget} summarizes the supervision configurations,
and Table~\ref{tab:ablations} reports their NAVSIM v1 results.
Both tables use paper experiment numbers and include Exp0--Exp9.

\paragraph{Gradient Paths and Candidate Pools.}
Exp0 uses trajectory GRPO and BC without video supervision.
In Exp1, video gradients update only VideoDiT; trajectory and BC
gradients still update the shared encoder. Exp2 also allows video
gradients to update shared parameters. Both use Q2 targets.
Exp3 shuffles rendered targets across trajectory conditions within a
scene after selection. Scenes without a valid target permutation
contribute no shuffled video supervision.

Exp4 selects the two highest- and two lowest-reward candidates from
the execution- and pose-valid pool. If this pool contains $n<4$
candidates, the high and low subsets contain $\lceil n/2\rceil$
and $\lfloor n/2\rfloor$ candidates, respectively. The subsets are
disjoint, with reward ties broken by sampling index.
Exp4 applies the four safety checks before training on Q2 and Q3.
Exp5 renders four safe high-reward candidates and trains on all valid
Q1 and Q2 candidates. Exp6 uses the same high/low pool as Exp4 but
trains on all valid rendered candidates, including individually
unsafe candidates. In Exp4 and Exp6, an unsafe candidate's positive
trajectory advantage is additionally truncated to zero:
$A_i\leftarrow\min(A_i,0)$. Exp2 does not apply this projection.
Exp7 increases the rendering budget from four to eight while retaining
Exp2's Q2 selection and video-training rule. Both settings train on all
valid Q2 candidates. Table~\ref{tab:video-budget} distinguishes the
rendered pool from the subset used for video training.

\paragraph{Fixed Offline Cache (Exp9).}
Exp9 follows the same experimental settings as Exp2, except that
trajectories are sampled offline from the initial model and their
rendered videos are stored in a fixed cache. The cached
trajectory-video pairs are reused throughout post-training.
All other settings, including candidate selection, rendering budget,
training objectives, and optimization hyperparameters, are identical
to Exp2. Exp9 achieves 90.4 PDMS, compared with 89.1 for Exp0
and 91.3 for Exp2 (Table~\ref{tab:online-offline}).

\begin{table}[t]
\centering\small
\caption{Video-supervision configurations. $K$ is the maximum number
of rendered candidates per observation in each sampling round.
For Exp9, this budget applies to offline cache construction.
FM denotes flow matching.}
\label{tab:video-budget}
\begin{tabular*}{\linewidth}{@{\extracolsep{\fill}}lclllr@{}}
\toprule
Setting & $K$ & Rendered pool & Video-training set & Video objective & PDMS\\
\midrule
Exp0 & 0 & None & None & None & 89.1\\
Exp1 & 4 & High reward & Q2 & FM & 89.1\\
Exp2 & 4 & High reward & Q2 & FM & 91.3\\
Exp3 & 4 & High reward & Q2, shuffled & FM & 83.1\\
Exp4 & 4 & 2 high + 2 low & Q2 + Q3 & FM & 90.9\\
Exp5 & 4 & High reward & Q1 + Q2 & FM & 91.0\\
Exp6 & 4 & 2 high + 2 low & Q1--Q4 & FM & 90.9\\
Exp7 & 8 & High reward & Q2 & FM & 91.2\\
Exp8 & 4 & High reward & Valid rendered candidates & Video GRPO & 91.0\\
Exp9 & 4 & Initial-policy cache & Q2 from cache & FM & 90.4\\
\bottomrule
\end{tabular*}
\end{table}

\begin{table}[t]
\centering\small
\caption{NAVSIM v1 ablation results. Configurations are listed
in Table~\ref{tab:video-budget}.}
\label{tab:ablations}
\begin{tabular*}{\linewidth}{@{\extracolsep{\fill}}lrrrrrr@{}}
\toprule
Setting & NC & DAC & TTC & Comfort & EP & PDMS\\
\midrule
Exp0 & 97.1 & 96.7 & 96.0 & 99.8 & 83.6 & 89.1\\
Exp1 & 97.6 & 96.7 & 96.4 & 99.7 & 83.3 & 89.1\\
Exp2 & 98.0 & 97.2 & 97.0 & 100.0 & 86.7 & 91.3\\
Exp3 & 94.1 & 91.3 & 91.8 & 99.6 & 81.2 & 83.1\\
Exp4 & 97.4 & 97.3 & 96.4 & 99.9 & 86.4 & 90.9\\
Exp5 & 98.0 & 97.1 & 97.0 & 99.9 & 86.1 & 91.0\\
Exp6 & 97.7 & 97.2 & 96.6 & 99.9 & 86.5 & 90.9\\
Exp7 & 97.8 & 97.3 & 96.9 & 99.9 & 86.6 & 91.2\\
Exp8 & 97.9 & 97.2 & 96.9 & 99.9 & 86.2 & 91.0\\
Exp9 & 97.1 & 97.2 & 95.9 & 99.9 & 85.9 & 90.4\\
\bottomrule
\end{tabular*}
\end{table}

\paragraph{Video GRPO Control.}
Exp8 evaluates video GRPO under the same four-candidate rendering
budget used by Exp2. Exp8 applies video GRPO to all valid rendered
high-reward candidates, whereas Exp2 first selects the Q2 subset and
optimizes the selected targets with flow matching.
For each candidate, it generates future latents with five stochastic
Euler steps and noise scale 0.3, using preceding target frames as context.
The reward is the negative framewise latent MSE, aggregated with
normalized temporal weights proportional to $0.85^{r-1}$.
It does not include the motion-weighted temporal-change term in
Eq.~\ref{eq:tf-error}. Different trajectory conditions within the
same scene form the video comparison group.
The video policy objective uses group-normalized advantages and a
clipped ratio computed from dimension-normalized Gaussian log
densities, with clipping parameter 0.2.
One future frame is replayed for optimization per update, cycling
through the video horizon. Groups with fewer than two valid candidates
or reward standard deviation below $10^{-6}$ contribute no video
policy update. Video GRPO replaces video flow matching and updates
VideoDiT and shared parameters; its reward does not enter trajectory
advantages.

\paragraph{Rendering and Optimization Budgets.}
$K$ specifies the number of candidates budgeted for rendering;
Table~\ref{tab:video-budget} separately lists those used for video
training. Error-based selection follows rendering and scoring, so
these costs apply to every rendered candidate. Total training cost
also includes trajectory sampling, PDM evaluation, video optimization,
and optimizer updates. Comparisons at equal $K$ match rendering
caps, while video optimization costs depend on the training subset
and objective. Exp9 renders its cache once; Exp2 renders candidates
at each sampling round.

\paragraph{Interpreting the Ablations.}
Table~\ref{tab:video-budget} summarizes the supervision configurations
underlying the main-text ablations. For Exp3, scenes without a valid
within-scene target permutation contribute no shuffled video supervision.
Exp2 and Exp5 compare Q2 versus Q1+Q2 supervision from the same
rendering pool, although the number of video-training candidates differs.
Exp4 and Exp6 additionally modify the safety-specific treatment of
trajectory advantages and are therefore interpreted as broader
configuration comparisons.

\paragraph{Results across Random Seeds.}
Table~\ref{tab:seed-results} reports NAVSIM v1 PDMS for Exp0, Exp2,
Exp5, and Exp7 across three random seeds. Exp2 outperforms the other
three variants on every seed.

\begin{table}[t]
\centering\small
\renewcommand{\arraystretch}{1.12}
\caption{NAVSIM v1 PDMS across random seeds. The final column reports
the mean and sample standard deviation over three runs.}
\label{tab:seed-results}
\begin{tabular*}{0.75\linewidth}{@{\extracolsep{\fill}}lcccc@{}}
\toprule
Setting & \multicolumn{3}{c}{Random seed} & Mean $\pm$ SD\\
\cmidrule(lr){2-4}
 & 1234 & 20260901 & 20260910 & \\
\midrule
Exp0 & 89.1 & 89.2 & 89.0 & $89.1 \pm 0.1$\\
\textbf{Exp2} & 91.3 & 91.4 & 91.3 & $91.3 \pm 0.1$\\
Exp5 & 91.0 & 90.9 & 91.0 & $91.0 \pm 0.1$\\
Exp7 & 91.1 & 91.2 & 91.2 & $91.2 \pm 0.1$\\
\bottomrule
\end{tabular*}
\end{table}

\section{Video Supervision and Evaluation}
\label{app:video}
The teacher‑forced error quantifies model prediction consistency against rendered simulation targets given preceding reference frames. We separately evaluate renderer fidelity, autoregressive video generation quality, and action‑conditioning accuracy.

\paragraph{Renderer Fidelity.}
We compare front-view World Engine renderings along logged GT
trajectories with real observations on NAVSIM v1 NavTest.
Each clip contains eight future frames over four seconds at 2 Hz.
FID is computed on frames and FVD on clips using the same preprocessing
and feature extractors for the two domains.
These measurements quantify rendering fidelity at logged viewpoints;
novel-view coverage is discussed in Appendix~\ref{app:limitations}.

\paragraph{Model Video Quality.}
For each evaluation sample, Exp0 and Exp2 generate autoregressive
video predictions along the same logged GT trajectory used for the
World Engine rendering. Both models use the same observation history.
We use the same dataset, clip length, preprocessing, and feature
extractors as the renderer evaluation. FID and FVD compare the
predictions with the corresponding real observations.

\paragraph{Qualitative Examples.}
Figure~\ref{fig:matched-renderings} shows a shared observation history,
a bird's-eye view of three alternative executed paths (A--C), and
World Engine renderings along each path at $+1$, $+2$, $+3$, and
$+4$ seconds.
Figure~\ref{fig:video-trajectory-alignment} compares World Engine
renderings with predictions from \method and the Epona initialization
at the same displayed future times.

\section{Cross-Benchmark Results}
\label{app:benchmarks}

\paragraph{Evaluation Metrics and Scope.}
PDMS combines collision and drivable-area terms with progress, TTC,
and comfort. EPDMS additionally evaluates driving direction,
traffic lights, lane keeping, and further comfort measures.
On HUGSIM, HD-Score combines safety and comfort over the rollout
and scales the result by RC.
Higher scores are better. We evaluate \method and the NAVSIM-adapted
Epona initialization on the full NAVSIM v2 NavTest set using the
NAVSIM v2.2 devkit with single-stage EPDMS evaluation, LQR tracking,
and non-reactive traffic agents.

\begin{figure}[htbp]
\centering
\includegraphics[width=.96\linewidth]{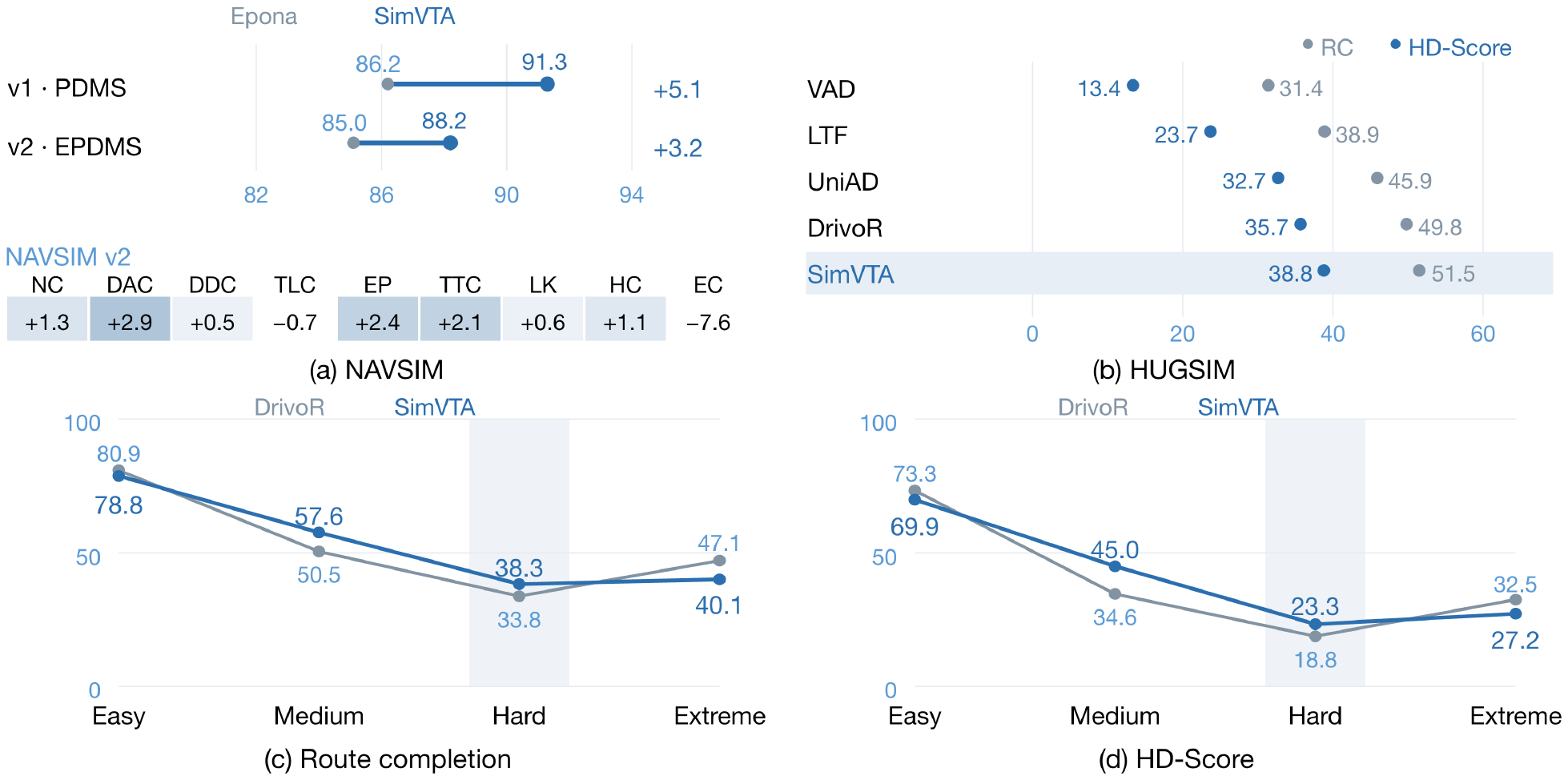}
\caption{\textbf{Reported planning results across benchmarks.}
(a) NAVSIM v1 PDMS and v2 EPDMS, with NAVSIM v2 component-score
differences (\method minus the NAVSIM-adapted Epona initialization).
(b) HUGSIM RC and HD-Score using benchmark aggregation.
(c) RC and (d) HD-Score for DrivoR and \method by scenario difficulty.
Values correspond to Tables~\ref{tab:main-planning},
\ref{tab:navsim-v2}, and \ref{tab:hugsim}.}
\label{fig:benchmarks}
\end{figure}

\paragraph{Result Sources and Comparison Scope.}
Figure~\ref{fig:benchmarks} summarizes the NAVSIM and HUGSIM results.
On NAVSIM v1, the full framework improves PDMS by 5.1 points over
the initialization (86.2 to 91.3). The comparison with Exp0
(89.1 to 91.3) measures the additional gain when video supervision
is added within the framework. NAVSIM v2 and HUGSIM extend the evaluation
to additional planning criteria and closed-loop driving.
On NAVSIM v2, the NAVSIM-adapted Epona initialization obtains 85.0 EPDMS
under the same evaluation protocol as \method, which reaches 88.2 (+3.2).
This aggregate improvement coexists with lower traffic-light
compliance (99.7 to 99.0) and extended comfort (67.6 to 60.0)
(Table~\ref{tab:navsim-v2}).
Baseline HUGSIM results in Table~\ref{tab:hugsim} are taken from
\citet{kirby2026drivor}.
\method achieves the highest reported averages among these methods
and leads on hard scenarios, while DrivoR has higher extreme-scenario
RC and HD-Score (47.1/32.5 versus 40.1/27.2).
These closed-loop results assess the final planner; isolating the
post-training gain on HUGSIM additionally requires the initialization
and Exp0 under the same routes, controller, and rollout settings.

\paragraph{Recent WAM Results.}
Table~\ref{tab:reported-wam-results} includes results from recent
WAMs discussed in the related work, including methods whose reports
also supply baseline values in the main tables.
These published results complement the within-backbone ablations;
the table retains each source's checkpoint and evaluation convention.

\begin{table}[t]
\centering
\small
\caption{Recent WAM results reported by their respective papers.
Training data, initialization, and evaluators differ across methods.}
\label{tab:reported-wam-results}

\vspace{4pt}
\begin{tabular}{@{}lcc@{}}
\toprule
Method & NAVSIM v1 PDMS & NAVSIM v2 EPDMS\\
\midrule
SimWAM \citep{zhao2026simwam} & 91.5 & 90.2\\
Latent-WAM \citep{wang2026latentwam} & -- & 89.3\\
WA-JEPA \citep{wang2026wajepa} & 91.8 & 91.7\\
\method & 91.3 & 88.2\\
\bottomrule
\end{tabular}
\par\vspace{5pt}
\noindent\parbox{\linewidth}{\footnotesize
Sources: SimWAM, arXiv:2608.07468v4, Tables 1--2;
Latent-WAM, arXiv:2603.24581v1, Table 1;
WA-JEPA, arXiv:2608.20974v2, Tables 1 and 3.
SimWAM's v1 and v2 results are from its post-RL and pre-RL checkpoints,
respectively. A dash indicates no result in the cited source.}
\end{table}

\section{Limitations}
\label{app:limitations}

\paragraph{Reconstruction Quality and Coverage.}
Occlusions and limited pose coverage can degrade renderings away
from logged paths, while renderer metrics assess only logged viewpoints.
Even with safety and pose‑support checks, rendering artifacts can still appear. A high prediction error can stem from either model limitations or poor‑quality simulation targets. Quality-aware filtering and evaluation at alternative poses
remain necessary.

\paragraph{Distribution Shifts.}
Current-policy trajectories broaden action coverage, but observation
histories remain log-based and rendered targets can differ from real
images. Reconstruction coverage and candidate filtering further limit
supervision. Teacher‑forced selection cannot capture autoregressive error accumulation. We leave robustness evaluation under shifted observation histories
and reactive traffic~\citep{jia2024bench2drive,jia2026bench2drivevl}
for future work.

\paragraph{Architecture and Training Cost.}
We validate our framework based on the Epona backbone. Generalization to alternative WAM architectures is left for future work. Online rendering, error scoring,
and video optimization increase training cost, while inference uses
direct trajectory prediction.

\end{document}